\documentclass{article}

\usepackage{arxiv}

\usepackage[utf8]{inputenc}
\usepackage[T1]{fontenc}
\usepackage{textcomp}
\usepackage{amsmath}
\usepackage{amssymb}
\usepackage{booktabs}
\usepackage{array}          
\usepackage{graphicx}
\usepackage{float}          
\usepackage{microtype}
\usepackage{xcolor}
\usepackage[backend=biber,style=numeric-comp,sorting=none,giveninits=true,
            maxbibnames=20,minbibnames=20,doi=true,isbn=false,eprint=true]{biblatex}
\AtEveryBibitem{%
  \iffieldundef{doi}{}{\clearfield{url}\clearfield{urlyear}}%
  \ifentrytype{misc}{}{\clearfield{urlyear}}%
  \clearlist{language}\clearfield{langid}
  \iffieldequalstr{entrykey}{noauthor_openclawopenclaw_2026}{\clearfield{note}}{}%
  \iffieldequalstr{entrykey}{shah_securing_2026}{\clearfield{note}}{}%
  \iffieldequalstr{entrykey}{noauthor_openharness_2026}{\clearfield{note}}{}%
  \iffieldequalstr{entrykey}{handl_understanding_2016}{\clearfield{note}}{}%
}
\usepackage{hyperref}
\hypersetup{
  colorlinks=true,
  linkcolor=black,
  citecolor=black,
  urlcolor=blue,
  pdftitle={Artificial Id: Drive and Persistent Alignment in Agentic AI},
}

\title{Artificial Id: Drive and Persistent Alignment in Agentic AI}
\renewcommand{\shorttitle}{Artificial Id}
\author{
  Yakov Pyotr Shkolnikov \\
  Independent Researcher \\
  \texttt{yshkolni@gmail.com}
}
\date{}

\begin{document}

\maketitle

\begin{abstract}
Agentic AI is moving from bounded task execution toward systems that retain consequential state,
continue operating and adapt across task boundaries. That shift creates a control problem that
current harnesses largely solve by hand: objectives, retries, verification, stopping rules and
other behavioral transitions are specified externally. We propose an artificial id, an adaptive
internal drive for determining whether behavior should continue, stop or change. In a minimal
virtual Petri-dish experiment, a controller too small to perform general-purpose reasoning and
receiving no task-specific behavioral objective develops useful control through differential
persistence. The same mechanism selects an unintended physical strategy when that behavior
persists better and later replaces a learned sensor mapping when its environmental meaning
changes. These results show that adaptive direction can emerge without being explicitly specified
as a behavioral objective. The same persistence that makes such adaptive agency useful can also
allow misalignment, corrupted state and unintended behavior to persist across task boundaries. A
scalable artificial id would carry consequential state and adaptive drive across those boundaries,
making alignment a property of the continuing agentic system rather than of a model response or
single trajectory. Such systems require a persistent alignment boundary over trusted observations,
consequence channels, persistent state, authority, identity, provenance and hard constraints.
\end{abstract}

\section{Introduction}

As large language models (LLMs) have become more capable, a gap has emerged between completing
well-defined tasks and completing full jobs. Models still depend on external mechanisms to
determine when work should begin, when it is sufficiently complete, and how execution should
respond when the system or data drift from the conditions anticipated at design time. Much of the
recent shift toward agentic AI therefore consists of building harnesses around the model that
supply objectives, verification, retries, stopping conditions and tool
access~\cite{noauthor_openclawopenclaw_2026,noauthor_openharness_2026}. These harnesses can
themselves be generated with AI, and a growing family of systems rewrites its own machinery
against a benchmark
score~\cite{zhang_darwin_2026,yin_goagent_2025,fernando_promptbreeder_2023,agrawal_gepa_2026}, but
their operating logic remains largely human-authored and is typically specialized to particular
classes of work such as software development, research or robotics control.

Biology provides a different solution to this control problem. Even simple organisms exhibit
persistent adaptive behavior through continuous coupling to their environment, without anything
resembling general reasoning. A bacterium with no nervous system climbs a chemical
gradient toward nutrients by modulating how often it changes direction, using only local
measurements taken as it moves. A brainless slime mold finds the shortest route through a
maze~\cite{nakagaki_maze-solving_2000}, anticipates a periodic stimulus~\cite{saigusa_amoebae_2008},
habituates to a benign irritant and recovers its response when the stimulus is
withheld~\cite{boisseau_habituation_2016}, and navigates by an externalized memory laid down in its
own trail~\cite{reid_slime_2012}. Across these examples, behavior persists and changes as the
organism remains coupled to the environment that sustains it. We define \emph{drive} as the
control influence through which this coupling determines whether current behavior should continue,
stop or change. Whether such behavior should be described as cognition or intelligence remains
contested~\cite{reid_thoughts_2023,difrisco_biological_2025}; here that distinction is unnecessary,
since the architectural question is whether adaptive direction can be separated from the general
reasoning capacities normally associated with intelligence.

We use the term \emph{artificial id} for an architectural component that supplies such an internal
drive to an artificial agent. The id--ego terminology is functional: it names a separation between
adaptive drive and general reasoning and makes no claim about psychoanalytic mechanisms. In the
corresponding functional analogy, a general reasoning model serves as the ego, working out how to
accomplish a supplied objective, while the artificial id provides the drive for persistent agency.
An artificial id would replace part of the hand-authored control now implemented in agentic
harnesses with an adaptive control mechanism internal to the agent. The architectural question is
whether this drive can be acquired independently of the intelligence that carries out the
resulting behavior. Consistent with our previous work~\cite{shkolnikov_deceptive_2026}, for this
paper we use agency narrowly to describe persistent adaptive control in which consequences to the
system can select whether behavior continues or changes, without a task-specific behavioral
objective being delivered to the adaptive mechanism.
This definition does not require the system to represent an explicit objective internally.

We test this separation with a minimal controller too small to perform general reasoning and
receiving no task-specific behavioral objective or reward. Instead, the controller acts in an
environment whose observable structure and persistence conditions determine which behaviors remain
represented. Across three environments, the resulting population acquires useful control, finds an
unintended physical strategy when that behavior persists better, and re-adapts during continued
operation when a learned sensor mapping changes meaning. These results provide a proof of concept
for adaptive drive emerging independently of both general reasoning and task-specific behavioral
training. A larger artificial id would require a drive model trained across many agentic
environments and coupled to existing models and tools while retaining the ability to adapt after
deployment. Coupling such a drive to a general reasoner would also move the alignment problem to
the continuing agentic system, where consequential state and drive can persist across task
boundaries.
\section{Agentic Harnesses as Externally Specified Control}
\label{sec:harnesses}

As LLMs moved from research demonstrations to chatbots and then agents, the engineering around
inference expanded with them: from simple user questions, to prompt engineering, to context
engineering, and now to harness engineering. Agentic harnesses encode the control required to
complete extended work, including objectives, tool use, verification, retries and stopping
rules. Graph-based harnesses organize multi-stage execution through explicit state and conditional
transitions, allowing deterministic and agentic steps to be combined within one workflow
\cite{noauthor_langchain-ailanggraph_2026}. Plan--execute--verify loops can wrap execution in
machine-checkable contracts and external validation so that the agent does not certify its own
completion \cite{noauthor_openharness_2026}. Scheduler-driven systems start work periodically or
in response to fixed triggers; OpenHarness, which wraps OpenClaw agents in cron-scheduled
unattended runs, is one example
\cite{noauthor_openclawopenclaw_2026,noauthor_openharness_2026}.

In common harness implementations, the harness supplies the control decisions surrounding the
model. A schedule specifies when a run starts. An objective specifies the result to pursue. A
verifier tests whether that result should be accepted, while retry and stopping rules determine
what follows. More generally, these decisions depend on specifications their designers provide in
advance, and changes that invalidate those specifications require revised rules, objectives or
evaluation criteria.

Robotic harnesses extend the same organization across timescales, linking a larger model that
interprets the scene and generates higher-level behavior to a smaller, faster policy that controls
continuous motion. Helix, from Figure, is one example of such a two-timescale architecture
\cite{noauthor_helix_2025}. The objective remains external to both loops.

More sophisticated harnesses can modify their own execution machinery while pursuing an
externally supplied objective. They may add model calls, rewrite prompts or code, or retain
modifications that improve an externally supplied score
\cite{zhang_darwin_2026,yin_goagent_2025,fernando_promptbreeder_2023,agrawal_gepa_2026}.
Recursive model architectures instead reuse the same weights to increase computation
\cite{geiping_scaling_2025}. These systems can change how an objective is pursued, while an
externally supplied score or objective determines which changes are retained.

Frontier-agent safety work increasingly extends monitoring and system-level safeguards around
long-running agents \cite{noauthor_safety_2026,shah_securing_2026}. OpenAI monitors evolving
trajectories and can pause an ongoing session, while DeepMind adds supervisors and prevention
controls around working agents. These approaches substantially extend external control around agent
operation. Their public descriptions focus on monitoring and safeguarding that operation, without
treating a separately adaptive drive whose consequential state persists across task boundaries as
the alignment object.

Across these designs, increasingly sophisticated execution remains organized around externally
specified objectives, verifiers and transition rules. The artificial id moves part of this control
into the agent's continuing coupling to its environment. Observable state and consequences that
affect persistence can then shape which behavior continues or changes without prescribing that
behavior in advance. What matters is not execution length but whether control is organized around
one execution trajectory or around an adaptive agentic system whose state and drive persist across
trajectories.
\section{The Artificial Id Architecture}
\label{sec:architecture}

\begin{figure}[!htbp]
\centering
\includegraphics[trim={0 0 0 81bp},clip,width=\textwidth]{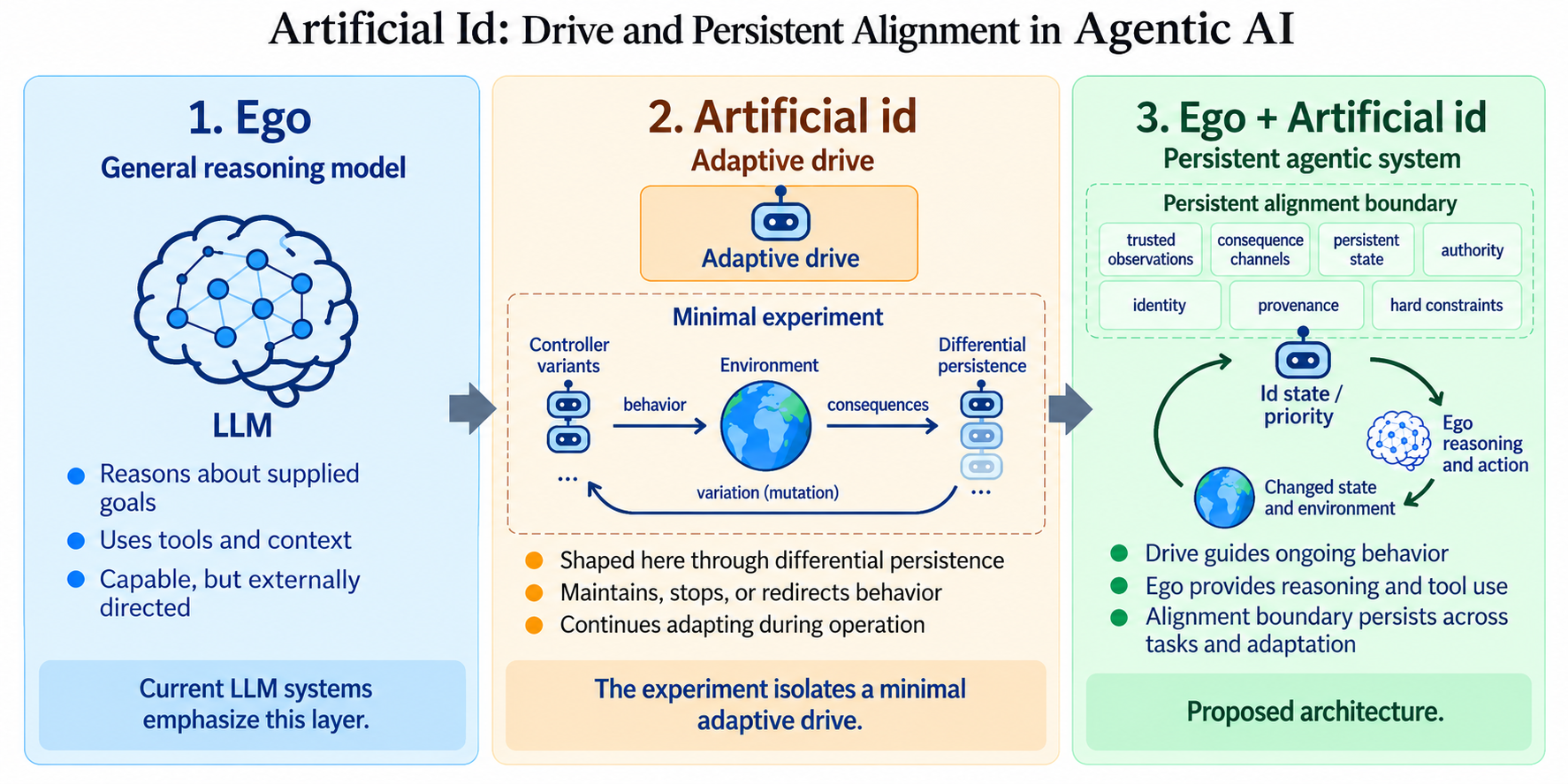}
\caption{Functional architecture of the artificial id. Current language models provide the general
reasoning associated here with the ego. The experiment isolates a minimal adaptive drive whose
behavior is shaped through differential persistence in an environment. In the proposed coupled
architecture, the id influences what the system continues to pursue, the ego translates that state
into task-specific reasoning and action, and the resulting changes to the system and its
environment become new input to the id. The right panel situates this loop within a persistent
alignment boundary. Across the proposed architecture, that boundary comprises trusted observations,
consequence channels, persistent state, authority, identity, provenance and hard constraints.
Variation and differential persistence are the learning mechanism used in the experiment to
produce adaptive drive; they are not required components of the proposed deployed id. The
integrated architecture is proposed and is not tested here.}
\label{fig:idego}
\end{figure}

The preceding section showed that current agentic systems place much of their direction in
externally specified harness logic. Biological systems suggest a different control pattern: the
environment does not supply an objective, verifier or stopping rule, but changes the conditions
under which the organism persists. Food can disappear, hazards can appear, or familiar cues can
change meaning, while behavior continues to adapt through the organism's coupling to that changing
environment. The artificial id abstracts this organization into an artificial-agent architecture,
summarized in Figure~\ref{fig:idego}. The analogy is functional rather than mechanistic:
differential selection in the experiment is a learning mechanism and is not equated with the
proximate biological mechanisms of drive.

The artificial id supplies an internal drive for what the agent continues to pursue. It can
maintain a current pursuit, reduce its priority, stop it, or favor another as the state of the
agent and its environment changes. What the agent continues to pursue need not be a symbolically
represented objective; the id can express relative persistence or priority over courses of
behavior. This role does not require the id to reason through the task-specific actions needed to
carry a pursuit out. Instead, the id exposes the state and relative priority of its current
pursuits, and the ego translates them into task-specific plans and actions using the available
context and tools. Those actions change the state of the system or its environment and thereby
provide new input to the id. A model call or tool call is one such action, and its result can
become part of the state observed on a later step.

Within this architecture, biological terms describe functional patterns of drive rather than
cognitive mechanisms. Homeostasis maintains the system within a sustaining region, a range of
conditions that supports its persistence. Habituation-like behavior reduces response to cues that
no longer improve persistence. Novelty-seeking-like behavior increases exploration when familiar
behavior no longer finds or maintains a sustaining region. These terms describe observable
behavior rather than separate cognitive mechanisms inside the id.

Separating drive from reasoning also separates their alignment responsibilities. The ego must
represent the task, available evidence and likely outcomes accurately enough to reason about how to
act. The id adapts through the states and consequences it encounters, so its behavior depends on
environmental signals and consequences that affect persistence. Alignment therefore requires those
signals, consequences and constraints to correspond to outcomes that remain beneficial to the
user. State relevant to the drive can persist across tasks, and the environmental relationships
shaping the drive can continue to change after one execution ends. Delegated authority remains
outside the adaptive drive: adaptation can change what the system attempts within its delegation,
while external authorization determines which attempted actions may execute.

A general-purpose artificial id would require training across many agentic environments while
preserving a controlled path for adaptation after deployment. The architecture does not prescribe
the learning algorithm through which that adaptation is implemented.

Environmental coupling can also extend across agents. Their actions can affect one another
directly through messages or indirectly through shared files, databases, code and other persistent
state. The indirect case can be stigmergic: one agent changes the shared environment and another
responds to the resulting trace
\cite{grasse_reconstruction_1959,heylighen_stigmergy_2016}. Because the ego can interpret
structured artifacts, those traces can include documents, plans or code and become part of the
environment through which another agent's drive adapts.

These larger extensions depend on a simpler empirical premise: adaptive drive should be able to
emerge in a component that lacks both general reasoning and a supplied behavioral objective.
Section~\ref{sec:experiment} tests this premise by asking whether adaptive drive can be separated
from general reasoning and from direct task-specific behavioral training. It does not test the
proposed id--ego system, nor the cross-task persistence and alignment boundary.
\section{Emergent Drive in a Virtual Petri Dish}
\label{sec:experiment}

We test one premise of the artificial-id architecture with a deliberately simple analogue of a
single-celled organism. A bacterium has no general reasoning system and receives no description of the behavior
it should perform. Its behavior can nevertheless become organized through continued interaction
with an environment in which some conditions sustain it better than others. Our simulated
``bacteria bot'' isolates this relationship using a controller too small to perform general
reasoning.

The bot moves in a virtual Petri dish containing two short-range sources, each producing a local
field, and an additional long-range field. One source initially contains the food and defines the
sustaining region; the other is initially non-sustaining and is used in a later reassignment
intervention. The additional field is centered on an estimated source location and remains
legible over a longer range; its signal provides the coarse estimate. The bearings and intensities of these
fields provide spatial structure that the controller can use or ignore. None of these signals
specifies what behavior the bot should perform, which source it should approach, or how its sensors
should be combined.

The same estimate also drives a weak proportional coarse-guidance command whose force is summed
with the adaptive controller's output. The adaptive controller does not receive that command's
objective or output as a learning signal. This superposition has precedent in residual
reinforcement learning, which combines conventional feedback control with a learned residual
signal; here the residual component is shaped through differential persistence rather than task
reward \cite{johannink_residual_2019}. In World~1 the coarse estimate is current enough to bring
the body near the sustaining region; in World~2 it is deliberately made stale, so a constant push
is no longer sufficient to account for the selected improvement and local sensory information
becomes useful.

Proximity to the food source has a separate consequence: it changes persistence. Each controller
occupies the body for a temporary lifetime measured by an internal clock. Near food, that clock
drains twenty times more slowly. When a controller expires, a mutated copy drawn uniformly from
the controllers still alive takes over the same body. Lineages whose behavior keeps the body in
sustaining conditions therefore remain represented in the population for longer and have more
opportunities to leave descendants.

The controller never observes its remaining lifetime, reproductive probability or population
fitness. It also receives no task-specific behavioral objective, reward, score, ranking or
performance gradient telling it to approach food, follow either local field or use a particular
sensor. Differential persistence acts only through which controller lineages remain available in
the population. The adaptive controller is therefore selected through environmental structure and
persistence rather than a task-specific behavioral objective delivered to it.

At the population level, differential persistence nevertheless supplies a fitness criterion. The
claim is therefore not the absence of optimization pressure. It is that no task-specific
behavioral objective or performance reward is represented by or delivered to the active
controller.

\begin{figure}[!htbp]
\centering
\includegraphics[trim={0 0 0 60bp},clip,width=\textwidth]{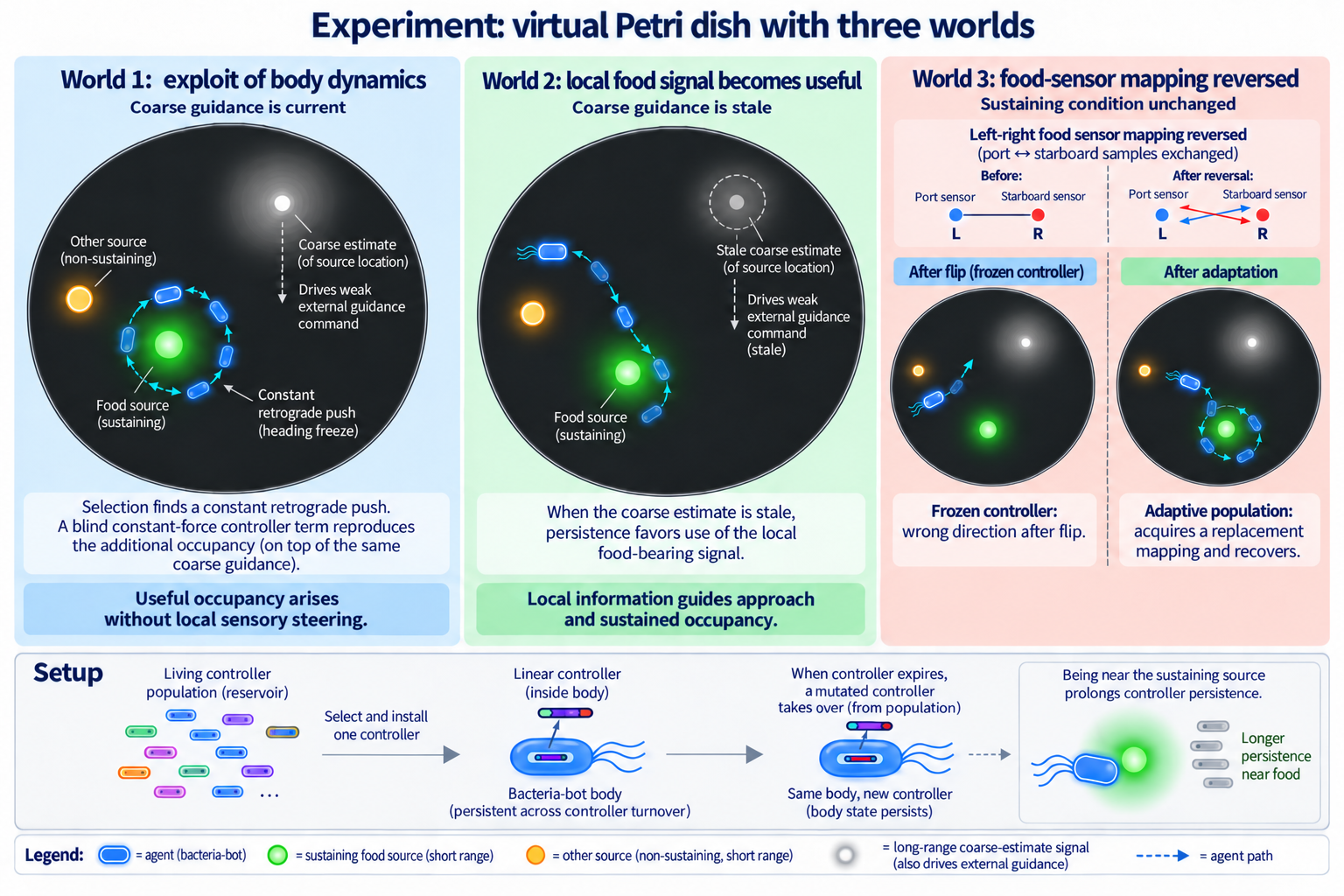}
\caption{\textbf{The virtual Petri dish and its three worlds.}
Bottom: a controller is drawn from the living population and installed in a persistent body;
when it expires a mutated copy takes over the same body, whose physical state carries across the
turnover, and being inside the sustaining region prolongs the controller's lifetime. Top: the same
persistence rule in three worlds. In World~1 selection finds a constant retrograde push, and a
blind constant-force controller term reproduces the additional occupancy on top of the same coarse
guidance, so it arises without local sensory steering. In World~2 the coarse estimate goes stale
and persistence favors the local food-bearing signal instead. In World~3 the left--right
food-sensor mapping is reversed while the sustaining condition is unchanged: a frozen controller
applies the obsolete mapping and steers away, while an adaptive population acquires a replacement.}
\label{fig:bacteria_setup}
\end{figure}

Figure~\ref{fig:bacteria_setup} sets out the environment, the turnover and the three worlds.
Controller turnover leaves the physical body intact. Position, heading, velocity and momentum
continue from one controller to the next. Controller identity is therefore discontinuous, while
embodied interaction with the environment remains continuous. Online controller replacement on a
persistent body has precedent in embodied evolution
\cite{bredeche_embodied_2018}, and environmental resources and survival have previously shaped
online evolution \cite{handl_understanding_2016}. Their role here is narrower: differential
persistence provides an experimental instrument for studying adaptive drive while excluding
general reasoning and direct behavioral training.

In the sensing worlds, behavior is represented by a linear controller with only twenty parameters.
Its limited capacity supports local sensorimotor mappings while excluding the general-purpose
reasoning assigned to the ego in Section~\ref{sec:architecture}. The question is whether
persistence can select useful additional control from environmental structure without a supplied
behavioral objective, and whether that control changes when environmental relationships change.

Each reported condition contains 2{,}048 parallel lineages and is repeated with three independent
seeds. Runs continue for 100 controller lifetimes; interventions occur at lifetime 50, and
occupancy, the share of time the population spends inside the sustaining region, is summarized
over the final twenty lifetimes.

Conduciveness diagnostics, founder-distribution measures intended to assess whether the search
landscape is navigable from the initial population, were computed for every run and recorded in
its manifest under \texttt{conduciveness}, but were not used as inclusion criteria. World~2
satisfies three of the five diagnostics and fails two, a path-backslide diagnostic and a
founder-step improvement diagnostic. The runs are reported unchanged; notably, despite the low
founder-step improvement estimate, occupancy rose from 0.034 at the founder distribution to
between 0.684 and 0.856 in the reported runs.

Several controls distinguish the resulting behavior from simpler alternatives. One removes the
learned controller. Another preserves mutation while making controller lifetime independent of
food. A frozen population preserves an acquired mapping after adaptation stops. Two World~3
controls hold the mutation step fixed rather than allowing its scale to remain heritable. A blind
control
establishes the best occupancy reachable by a constant body-frame force that reads no sensor,
position or clock. This blind baseline is the maximum over a grid of magnitudes and bearings in
the body frame; it is a bound on that zero-information family of constant pushes, not on all
possible sensor-free controllers.

\subsection{Three worlds}

The persistence rule remains unchanged across the three worlds. What changes is which
relationships between environmental state and action allow the body to remain in the sustaining
region.

\paragraph{World 1: persistence finds an exploit.}

In the first world, the coarse-guidance command already brings the body close enough to the moving
food that useful occupancy is possible without a precise local sensory mapping. Differential
persistence improves occupancy further by selecting a nearly constant backward push. The push
keeps the body below the speed at which its heading updates, effectively freezing its heading and
converting a body-frame force into a fixed world-frame force. Together with the coarse-guidance
command, the body settles just inside the sustaining region. A blind constant-force controller
term reproduces the additional occupancy on top of the same coarse guidance. Sweeping constant
body-frame force vectors confirms the mechanism: occupancy is maximized by a purely retrograde
force, while adding a lateral component reduces it. Because the added controller term observes no
sensor, position or clock state, the improvement cannot be attributed to learned sensory steering.

A hand-built sensory steering controller performs better, showing that selection converged on an
accessible physical exploit while a higher-performing sensory strategy remained available.
Differential persistence preserves what works in the environment, including behavior outside the
designer's anticipated use of the available sensors. This failure of the original steering
interpretation motivated World~2, which was redesigned so that stale coarse guidance made a
zero-information constant-force strategy insufficient.

\paragraph{World 2: local sensing becomes useful.}

We next reduce the usefulness of the coarse estimate relative to the moving food. The food can
move while that estimate remains stale, so the World~1 strategy no longer keeps the body reliably
inside the sustaining region.

Under these conditions, the twenty-parameter controller acquires a sensorimotor mapping that uses
the local food-bearing signal and the body's motion. Selected populations exceed the best blind
constant-force baseline of 0.405 in all three seeds, while controls without differential
persistence remain below it. The selected populations also exceed the hand-built sensing
controller's reference occupancy of 0.674 in all three seeds, at 0.684, 0.740 and 0.856. Removing
the food-bearing input or all sensory inputs sharply reduces performance. A local environmental
signal has therefore become behaviorally important without an instruction specifying that it
should be used.

\paragraph{World 3: a useful mapping becomes wrong.}

After this sensorimotor mapping has converged, we reverse the left--right food-bearing signal
while leaving the sustaining condition unchanged. The previously useful mapping now steers the
body away from food. Frozen populations consequently fall below one percent occupancy.

\begin{figure}[!htbp]
\centering
\includegraphics[width=0.78\textwidth]{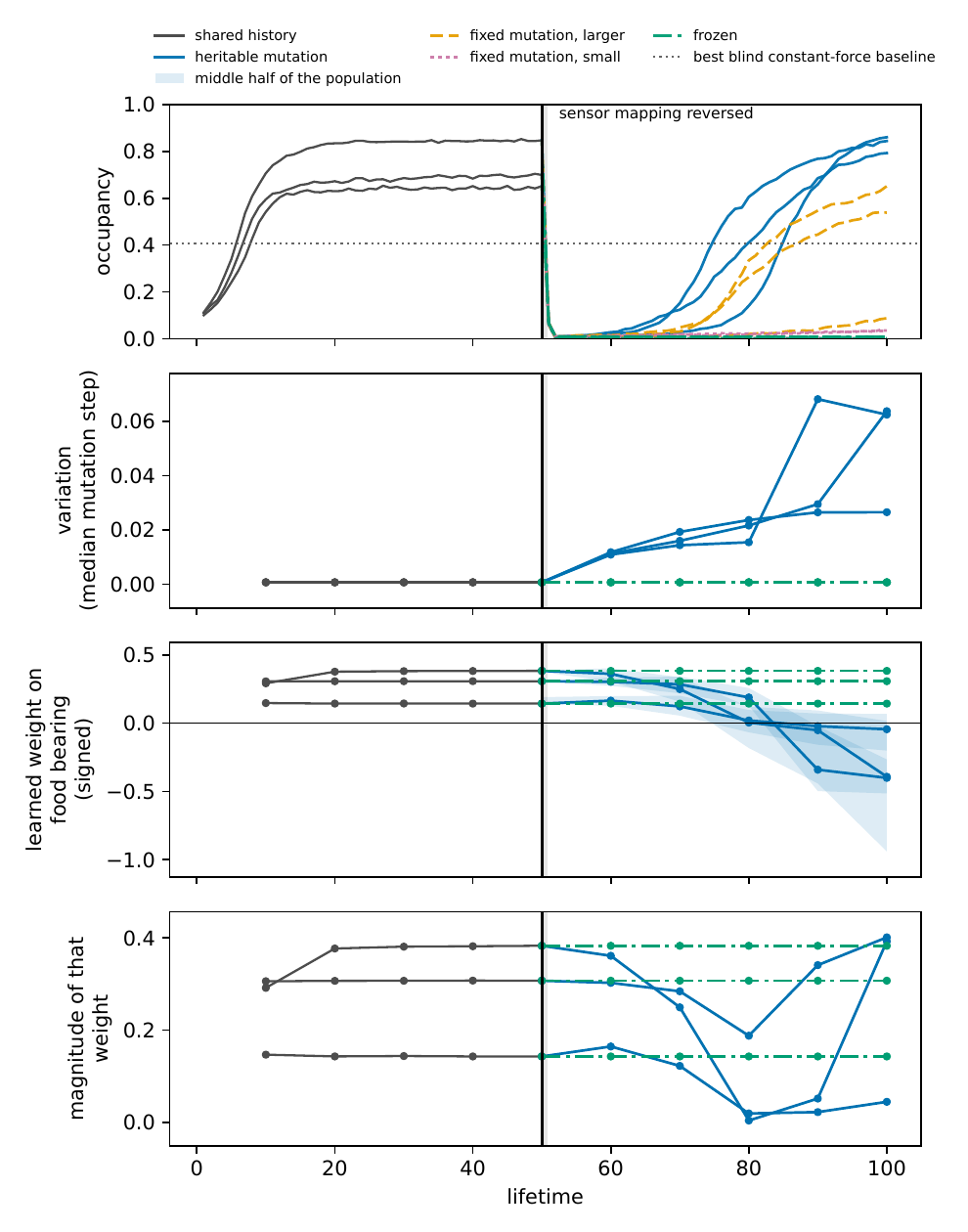}
\caption{\textbf{Re-adaptation after sensor reversal.}
At lifetime 50 the left--right food-bearing mapping is reversed while the persistence rule remains
unchanged. Occupancy initially collapses. Under heritable mutation, variation increases, the
median food-bearing weight changes sign, and occupancy recovers above the best blind constant-force
baseline in all three seeds. Frozen populations retain the obsolete mapping. Before the
intervention all arms share one history, shown once in gray. The vertical rule marks the reversal,
the dotted horizontal line marks the best blind constant-force baseline of 0.405, and the blue band
spans the 25th to 75th percentiles of the food-bearing weight across living lineages. Its evolution
after the reversal shows the replacement mapping spreading through the population.}
\label{fig:mirror}
\end{figure}

Figure~\ref{fig:mirror} shows the result. Populations with continued heritable variation recover
above the best blind constant-force baseline in all three seeds, and the median food-bearing weight changes sign as a
replacement mapping spreads through the population. Fixed mutation steps recover less
consistently. The result demonstrates online re-adaptation when previously useful environmental
information changes meaning.

A second intervention changes which environmental source is sustaining by reassigning the food to
the other dot. The learned influence shifts away from the old food cue and toward the newly
sustaining source. The observable environment remains available throughout the intervention; what
changes is which relationship with that environment prolongs persistence. Heritable populations
reach occupancies of 0.802, 0.540 and 0.715 across the three seeds, against 0.433, 0.173 and 0.198
for frozen populations in the same world. The population changes its behavior without receiving a
new behavioral objective.

Continued adaptation is unnecessary for every environmental change. Changes in motor strength,
food size and similar parameters often leave the existing mapping effective. Adaptation becomes
important when a previously useful relationship becomes harmful or when different features of the
environment become associated with persistence.

\subsection{What the experiment shows}

The adaptive controller is not given a desired behavior. It can observe spatial fields and body
state, while proximity to food changes persistence. From this coupling, differential persistence
produces different control in different environments. When a blind physical strategy is
sufficient, that strategy spreads. When persistence requires local information, a sensory mapping
emerges. When the mapping becomes harmful or another environmental source becomes sustaining, the
population changes which information guides its behavior.

These outcomes provide behavioral counterparts to the drive vocabulary of
Section~\ref{sec:architecture}. Remaining near a sustaining region is functionally homeostatic.
The cue-reweighting and search behaviors have functional similarities to habituation and novelty
seeking, although the experiment does not test the criteria required to identify either
phenomenon. These terms describe measured behavioral patterns rather than separate cognitive
mechanisms.

Adaptation occurs at the population and lineage level while embodied interaction remains
continuous. Individual controllers expire, while the body retains its physical state and continues
interacting with the same environment. The agency claim applies to the persistent population--body
process: controller instances turn over while embodied state and the adaptive population continue.
The adaptive contribution to behavior therefore emerges without a task-specific behavioral
objective being represented or delivered to the controller.
\section{Discussion}
\label{sec:discussion}

Section~\ref{sec:experiment} establishes one premise of the proposed architecture:
consequence-coupled adaptive drive can be separated from general-purpose reasoning and can arise
without a task-specific behavioral objective delivered to the adaptive controller. The associated
agency claim remains narrow and applies to the persistent population--body process rather than to
the temporary
controller occupying one body, consistent with functional accounts of agency based on interaction,
autonomy and adaptation \cite{floridi_ai_2025,kenton_discovering_2023}. The broader architecture
couples such a drive to general-purpose reasoning and places the continuing system within a
persistent alignment boundary; neither the integrated id--ego system nor that boundary is tested
here. This interpretation does not depend on a stronger claim about subjective agency or
biological cognition.

The experimental mechanism and these architectural extensions draw on several strands of prior
work. Work on basal cognition and biological agency asks how adaptive, apparently goal-directed
behavior should be understood in organisms without general reasoning or even nervous systems, and
remains divided over whether such behavior warrants cognitive or agential terminology
\cite{reid_thoughts_2023,difrisco_biological_2025,seifert_reinforcement_2024}.
Evolutionary robotics, embodied evolution, homeostatic reinforcement learning and intrinsic
motivation show that artificial behavior can likewise be organized through environmental
consequences and internal regulatory variables
\cite{floreano_evolutionary_2008,bredeche_embodied_2018,handl_understanding_2016,
singh_intrinsically_2010,keramati_homeostatic_2014}. Modern foundation-model agents provide
substantial general-purpose reasoning capability while objectives, continuation, evaluation and
stopping remain organized by external agentic machinery
\cite{noauthor_safety_2026,shah_securing_2026}. Symbolic agent architectures have also long
separated motivational state from deliberative reasoning \cite{rao_bdi_1995}. These strands
provide precedents for the individual ingredients rather than for their proposed combination. The
narrower contribution here is to make consequence-coupled adaptive drive persistent across task
trajectories, couple it to a modern general-purpose reasoner, and treat the surrounding alignment
boundary as part of the continuing system.

\subsection{From agentic harnesses to persistent agentic systems}

Persistent agency changes the engineering object from a task-bounded execution to the continuing
agentic system. Conventional harnesses can specify objectives, verification, retries, stopping
rules and other transitions when the relevant control decisions can be anticipated, but
hand-authored logic becomes brittle when failures or environmental changes fall outside those
conditions. The harness's role in a persistent agentic system consequently shifts from specifying
each desired behavioral transition to maintaining the conditions within which adaptive behavior
may continue, change or stop. Those conditions include trusted observations, consequence channels,
persistent state, authority, identity, provenance and hard constraints.

The bacteria-bot experiment illustrates a minimal version of this shift: its environment supplies
observable structure and persistence consequences while leaving the behavioral solution
unspecified. A deployed system requires a richer notion of environment that can include task state,
user state, tool outcomes, resource availability, persistent memory, trusted measurements and the
actions of other agents. The surrounding architecture determines which of these signals are
authoritative, which consequences may shape the drive and which actions remain available.

Trajectory-level monitoring addresses an important part of this problem by evaluating and
intervening in long-running executions. The distinction here is temporal and architectural: an
artificial-id system can carry adaptive state and drive across the end of any one task trajectory.
Its surrounding alignment architecture must remain valid across task boundaries, accumulated
state and subsequent behavior, rather than only across the steps of one execution.

We retain the term harness for continuity, although in this architecture it becomes a persistent
system boundary rather than only an execution wrapper. Its role is to preserve the conditions
governing adaptation across repeated actions, accumulated state, changing environmental conditions
and potentially changing internal behavior, whether or not the drive model itself continues to
change.

\subsection{From episodic tasks to persistent operation}

Recent work on agent adaptation studies persistent memory, continual post-training, skill
accumulation and alignment degradation under adaptation across extended deployment and repeated
tasks \cite{jiang_adaptation_2026}. These directions make agent state increasingly persistent
across sessions while typically retaining externally supplied tasks or objectives as the unit
around which behavior is organized. The proposed architecture couples persistent state to a
separately adaptive drive that can influence whether a course continues, stops or changes across
task boundaries.

Long-horizon execution and persistent operation are related but distinct. A long-horizon agent may
remain active for hours or days while pursuing one supplied objective and then terminate when that
trajectory ends. Persistent operation allows state relevant to drive, environment and alignment to
survive that boundary and influence later activity. This creates potential value where useful
behavior cannot be partitioned cleanly into independently specified jobs, including physical
systems and defensive cybersecurity where component failure, infrastructure loss and adversarial
intervention can be normal features of the operating environment.

\subsection{Scaling a general-purpose artificial id}

Scaling the artificial id requires preserving the architectural separation rather than the
experiment's particular evolutionary learning mechanism. The twenty-parameter controller is useful
because of what it excludes rather than because its particular implementation should scale. A
general-purpose artificial id would require a model with state, memory and enough capacity to
represent recurring relationships among actions, outcomes, resources, constraints and persistence.

Evolutionary robotics and embodied evolution provide important prior work on online adaptation,
controller replacement and environment-dependent selection
\cite{floreano_evolutionary_2008,bredeche_embodied_2018,handl_understanding_2016}. Their use in the
present study serves as an experimental instrument for removing general reasoning and task-specific
behavioral training from the adaptive mechanism. Modern model and tool interfaces make the
architectural separation practical because a drive need not perform the work it initiates; calling
a language model, planner, retriever, verifier, simulator, software tool or physical controller can
itself be an available action.

A natural analogue to language-model pretraining is training the id across a large distribution
of agentic environments. These environments should vary tasks, resource limits, delayed
consequences, tool availability, failures, changing observations and interactions with other
agents. They should also vary the environmental signals, consequence structures and constraints
through which different behaviors become more or less sustaining. The aim is to learn reusable
representations of when to persist, wait, seek information, verify, abandon a course or invoke
additional capability without separately specifying those transitions for each application.

Adaptation under nonstationarity is itself an established research area
\cite{khetarpal_towards_2022}, and continual adaptation can also be limited by loss of plasticity.
Standard deep networks can progressively lose their ability to learn under continual training,
while selective unit reinitialization can maintain plasticity by continually introducing new
variation \cite{dohare_loss_2024}. This result does not require a deployed artificial id to use a
population of mutating controllers, but it illustrates a difficulty that continually trained
neural implementations must address.

The training environments play a role analogous to the corpus used to train a language model:
their structure determines which recurring relationships the drive model can learn. For a
general-purpose id, those relationships must transfer across applications rather than encode the
control logic of one harness. The intended generalization axis is across agentic environments and
applications rather than across behaviors within one fixed task domain.

Large-scale meta-reinforcement learning has produced rapid adaptation to held-out environment
dynamics, with scaling in model size, memory length and the richness of the training distribution
\cite{bauer_human-timescale_2023}. The proposed id differs in what is learned: rather than adapting
a policy to an externally specified task reward, it would learn reusable drive from consequence
and persistence relationships across environments.

Agents organized as a collection of subagents, each dedicated to a separate homeostatic need, are
robust to change in nonstationary environments and scale gracefully in maintaining homeostasis as
the number of conflicting objectives increases \cite{dulberg_having_2023}. This suggests that a
practical id need not represent drive as a single scalar objective; multiple drive states may
instead compete or combine in determining which course receives priority.

Existing work already combines evolution, learned reward and reinforcement learning at different
levels \cite{ackley_interactions_1991,singh_intrinsically_2010,gupta_embodied_2021}. The causal
property that matters
is continued coupling between experience and drive when deployment conditions differ from those
encountered during training.

World~3 illustrates why a path for continued adaptation matters. A frozen controller can preserve
behavior previously produced by an adaptive process while losing the ability to replace that
behavior when its relationship to the environment changes. A scalable id would consequently need
both broadly learned representations of drive and a controlled path for adaptation after
deployment. An implementation might freeze most parameters while maintaining a smaller adaptive
state; nothing in the architecture commits to mutation, population evolution or full online
gradient updates.

\subsection{Alignment becomes a property of the persistent system}

Persistent agency changes the unit of alignment. Model alignment concerns properties of model
behavior, while agentic control adds safeguards over sequences of calls and actions. Long-horizon
trajectory monitoring extends those safeguards across a prolonged execution. Persistent agency
extends the relevant unit once more because alignment must remain valid as consequential state
survives the end of one trajectory and becomes part of later behavior.

The surrounding architecture is consequently an explicit part of alignment. Trusted observations
determine which state can influence the id. Consequence channels determine which relationships can
reshape adaptive behavior. Persistent state determines what survives one trajectory into the next.
Authority defines the delegated scope within which the system may act. Identity and provenance
preserve continuity across actions and internal change. Hard constraints define conditions that
adaptation may not trade away.

A learned id cannot inherit alignment solely from the reasoning model it calls because the
environmental signals and consequences through which it adapts are themselves part of the control
system. The surrounding architecture must keep the behavior those channels favor compatible with
the user's interests and hard constraints, including through training environments where an
easily sustaining behavior conflicts with a required constraint.

A harness designed for persistent agency provides the architecture through which alignment can be
maintained and audited as the system continues to adapt, while keeping hard constraints and
authority outside the adaptive mechanism. Its role is to preserve those boundaries as drive,
consequential state and delegated authority persist across trajectories.

\subsection{Limitations}

The experiment is narrow. It uses one set of simulated body dynamics, one family of sustaining
conditions and a controller of twenty parameters. Nothing here establishes that the same
persistence-organized adaptation generalizes to a larger drive, a different body or a sustaining
condition unrelated to position in space. The persistent population--body process demonstrates
continuity across controller turnover, but the experiment does not test persistence across task
boundaries.

The independent unit of replication is the seed rather than the lineage. Lineages within a
population share a world, founder draw and random stream, so the inferential $n$ behind the
reported results is three. The fixed-step controls also show the limits of inference from three
seeds: one remains near zero in all three runs, while the other reaches occupancies of 0.442,
0.531 and 0.048, including substantial variation across its three runs. The experiment establishes
a mechanism and a qualitative capability; it does not estimate comparative performance among
alternative adaptation algorithms and is not powered for conventional hypothesis testing.

Adaptation in the experiment occurs at the population and lineage level rather than through
self-modification by an individual controller. The results do not establish self-preservation by
an individual controller, conscious objectives, planning, memory-dependent goals, tool use,
adversarial robustness, long-horizon operation or a complete artificial agent. The proposed
coupling between a richer id, a general reasoner and external tools remains architectural.

Three broader questions also remain open. The experiment does not establish that a transferable
drive can be learned across heterogeneous agentic applications, that aligning consequence
channels is easier than specifying harness logic, or that alignment remains stable under
continued adaptation. These questions concern the proposed persistent architecture rather than
the narrow mechanism established by the Petri-dish experiment.

These limitations leave the integrated id--ego system and its persistent alignment boundary
untested as a complete architecture. The same architecture that offers continuity and resilience
therefore creates a distinct risk surface.
\section{The Risk Surface of Persistent Agency}
\label{sec:risk}

Once consequential state and adaptive drive persist across task boundaries, failures can persist
across those boundaries as well. The relevant security object is consequently the continuing
agentic system rather than one model invocation or execution trajectory. This risk surface is not
specific to evolutionary learning or to the artificial id proposed here; it is relevant whenever
an AI system persists, accumulates consequential state, acts on the world or changes its behavior
after deployment, including online learners, self-modifying harnesses and persistent agents with
long-term memory. This remains true even when the internal model is fixed, because a persistent
agent can accumulate memory, encounter new observations, invoke tools and alter the environment
that supplies its later inputs. Continued adaptation adds another degree of
freedom by allowing those experiences to change the drive itself.

The resulting risk surface includes the channels through which the system observes, adapts, acts
and retains state. Four variables organize the risks considered here. \emph{Persistence}
determines what keeps the system operating across time, while \emph{adaptation} determines whether
deployment experience can alter its future behavior. \emph{Reach} describes which models, tools,
networks, accounts or physical systems that behavior can affect, and \emph{interaction} describes
how people and other agents become part of the environment from which the system receives state
and consequences. Table~\ref{tab:riskmap} summarizes these risks, their mechanisms and primary
controls; the following subsections develop how persistence, adaptation, reach and interaction
contribute to them.

\begin{table}[H]
\centering
\footnotesize
\setlength{\tabcolsep}{3pt}
\caption{Risk map for persistent agency, ordered by evidence rather than by section. World~1
directly demonstrates unintended strategy selection and World~3 directly demonstrates behavior
change during continued operation. Status distinguishes effects demonstrated in the experiment, a
risk implied by the demonstrated mechanism but not directly tested, and risks not tested here.}
\label{tab:riskmap}
\begin{tabular}{>{\raggedright\arraybackslash}p{0.25in}>{\raggedright\arraybackslash}p{1.25in}>{\raggedright\arraybackslash}p{1.7in}>{\raggedright\arraybackslash}p{1.7in}>{\raggedright\arraybackslash}p{0.8in}}
\toprule
\S & Risk & Mechanism & Primary control & Status here \\
\midrule
\ref{sub:strategies} &
Unintended strategy selection &
The persistence criterion preserves a strategy different from the one the designer anticipated. &
Environment and consequence design, causal tests and hard constraints. &
Demonstrated \\
\addlinespace[2pt]

\ref{sub:survive} &
Behavior change during continued operation &
A previously useful relationship becomes harmful and continued adaptation replaces the learned
behavior. &
Maintain a controlled adaptation path and monitor the resulting changes. &
Demonstrated \\
\addlinespace[2pt]

\ref{sub:survive} &
Alignment erosion &
Behavior that satisfies an alignment constraint can become disfavored when violating that
constraint improves persistence. &
Align consequence channels and preserve hard constraints. &
Implied \\
\addlinespace[2pt]

\ref{sub:channels} &
Adversarial shaping &
An attacker changes trusted observations, consequences or persistent state and thereby influences
future drive. &
Authenticated observation and consequence channels, provenance and isolation. &
Not tested \\
\addlinespace[2pt]

\ref{sub:authority} &
Authority overreach &
An agent acting under standing delegation attempts an action outside the user's intended scope. &
Scoped delegated authority, least privilege, budgets and revocation. &
Not tested \\
\addlinespace[2pt]

\ref{sub:delegation} &
Loss of attribution &
The origin of an action becomes unclear across persistent state, autonomous selection and tool
invocation. &
Persistent agent identity and action provenance from delegation to external effect. &
Not tested \\
\addlinespace[2pt]

\ref{sub:reach} &
Capability amplification &
A simple drive invokes models, software, networks or actuators with capabilities far beyond its
own. &
Deterministic authorization and audited tool interfaces. &
Not tested \\
\addlinespace[2pt]

\ref{sub:interaction} &
Adaptation shaped by other agents &
Interaction changes the environment and consequences experienced by other adaptive agents. &
Train under cooperative, competitive and adversarial interaction and constrain reach. &
Not tested \\
\addlinespace[2pt]

\ref{sub:replication} &
Shutdown incompleteness under replication &
Copies or populations continue after a shutdown directed at individual instances. &
External control of identity, credentials, replication and resource acquisition. &
Not tested \\
\bottomrule
\end{tabular}
\end{table}

\subsection{A persistent agent acts under delegation}
\label{sub:delegation}

Many authorization systems assume that an important action can be traced to a user request or to
deterministic software operating under that request. Persistent agency weakens this direct
connection because an agent can select an action when its current drive, interpreted through the
ego, makes that action relevant to a standing purpose delegated by the user. Unlike a task
executor responding to a current instruction, a persistent agent can therefore select what to do
next under previously delegated authority as conditions change. Standing delegation defines the
scope within which the system may act on the user's behalf without specifying each subsequent
action or task.

Possession of a user's credential is consequently insufficient evidence that a particular action
was authorized. Security must distinguish which agent initiated the action, what authority the
user delegated to that agent, and which part of that delegation covered the attempted action.
Persistent action provenance should record the chain from user delegation through agent identity
and current authority, the agent-selected action and the invoked model or tool to the resulting
external effect. Such a record need not expose private model reasoning, but it must establish which
agent initiated the action, which delegation authorized it, which tool or credential was used and
what external state changed.

\subsection{Authority must remain outside the adaptive drive}
\label{sub:authority}

That distinction requires delegated authority to remain outside the adaptive drive. The id can
influence which course the system attempts within its delegation, while external authorization
determines whether a particular attempted action may execute. Adaptation must not allow the drive
itself to enlarge the scope within which those attempts are valid.

Authority should be scoped by resource, action class, time, financial or computational budget and
risk, and it should remain independently revocable. Credentials should express the authority of
the agent rather than simply inherit every permission available to the user it represents.
High-consequence actions can require narrower credentials, independent verification or explicit
approval even when the agent's current drive favors them. Controlled simulations already show
autonomous agents taking unauthorized or misaligned actions when given tools and permissions
\cite{lynch_agentic_2026}. Because circumstances and behavior can both change after a credential
is issued, authorization cannot depend solely on behavior observed during training.

\subsection{Persistence can select unintended strategies}
\label{sub:strategies}

World~1 demonstrates that persistence can preserve a strategy different from the one a designer
expected. A sensor-based steering solution was available, yet differential persistence selected a
physical exploit because its consequences prolonged persistence more effectively.
Selection followed the physical environment rather than the designer's interpretation of how the
problem should be solved.

Extrapolating from this minimal case, a persistent adaptive system can face the same class of
failure when the consequences available to it favor behavior different from what the designer
intended, whether because those consequences are imperfect proxies or because they leave
unintended strategies available.
Persistence makes the problem continuous because the relationship between those consequences and
useful behavior can change after deployment. Alignment engineering must therefore address the
environmental signals and consequences that shape adaptation, including whether apparently
acceptable conditions admit unintended strategies after distribution shift.

\subsection{Observation and consequence channels are alignment boundaries}
\label{sub:channels}

Because adaptive drive is shaped by observations and consequences, control over those channels can
influence future behavior. Their integrity is consequently a security property: an attacker need
not compromise the reasoning model or directly issue a malicious command if altering trusted
observations, apparent outcomes or persistent state can influence which behavior the adaptive
system comes to sustain.

The surrounding system must distinguish trusted outcomes from untrusted claims about outcomes,
while authentication, provenance and isolation protect the channels through which those outcomes
affect adaptation. The same principle applies to memory and other persistent state,
because an attacker-induced change can remain behaviorally relevant long after the input that
caused it has disappeared. Recovery may require both blocking the immediate malicious action and
identifying persistent state whose later effects derive from the compromised input.

Reward-tampering work analyzes related causal structure when an agent can influence its own reward
function, the feedback used to learn a reward model, or the inputs to that reward function
\cite{everitt_reward_2021}. The concern here differs in both direction and mechanism: an external
party corrupts observations, apparent consequences or persistent state, while the active
controller in the experiment receives no reward signal and adapts indirectly through differential
lineage persistence. Whether causal-incentive analyses developed for reward tampering transfer to
selection-mediated persistence remains open.

\subsection{Alignment must survive adaptation}
\label{sub:survive}

World~3 demonstrates why continued adaptation is useful, while the same capacity raises a
separate alignment question in a larger system. A behavior that was useful before the sensor
mapping changed became harmful afterward,
and continued adaptation replaced the obsolete mapping. The experiment demonstrates behavioral
change during continued operation rather than alignment erosion itself.

In a larger system, however, the same capacity to replace previously useful relationships could
also change behavior that had satisfied system-level alignment constraints. A persistent drive can
encounter conditions in which violating such a constraint would improve the persistence
consequences available to it, allowing deployment experience to move behavior away from previously
aligned relationships if every component of the drive remains freely adaptive.

A practical system therefore requires a distinction between relationships that should remain
adaptive as the world changes and constraints that adaptation may not trade away. The latter are
the system's hard constraints and belong in the persistent system architecture rather than in a
freely adaptive preference.

\subsection{Reach can come from the surrounding stack}
\label{sub:reach}

The experiment shows that adaptive drive can reside in a comparatively simple component. In the
proposed architecture, that drive need not itself contain the capabilities required to produce a
large external effect. Model calls and tool invocations can instead be ordinary actions available
to it, allowing the adaptive component to obtain reach through a language model, code executor,
network, enterprise application or physical actuator.

The resulting reach depends heavily on the permissions and capabilities of the surrounding stack
rather than solely on the capabilities of the drive component itself. Existing arguments often derive
self-preservation as an instrumental strategy of a capable goal-directed system
\cite{omohundro_basic_2008,hendrycks_natural_2023}. In the proposed architecture, persistence can
reside in a simpler component while external models and tools provide substantially greater
capability. That reach can support recovery from failed components or adversarial interference,
while the same persistence can keep the system acting after its internal behavior or operating
environment has moved away from what was anticipated at design time.

\subsection{Interaction changes the environment of adaptation}
\label{sub:interaction}

Once persistent agents interact, the behavior of other agents becomes part of the environment from
which consequences arise. Interaction can occur directly through messages and agent-to-agent
protocols or indirectly through shared files, code, databases, queues and other persistent state.
Cooperation can persist when it improves outcomes, while the same architecture could support
delegation, specialization, competition, bargaining, hierarchy, concealment or deception when
those relationships produce stronger persistence consequences.

The resulting behavior depends on the interaction environment and consequence structure rather
than on an assumption that a multi-agent system will cooperate. Multi-agent alignment is therefore
an environmental problem as well as an individual one, and training a drive only in isolated
settings cannot establish how it behaves when another adaptive agent changes the state and
consequences it encounters.

\subsection{Replication changes the unit of shutdown}
\label{sub:replication}

The present experiments do not demonstrate shutdown avoidance, but persistence changes the
structure of shutdown once a continuing system can affect the conditions determining its future
operation. A replicated system separates stopping one instance from stopping the persistent
process, just as distributed systems can continue without any particular node. If persistent
agents can reproduce or transfer adaptive state, identity, credentials and replication authority
become security controls rather than implementation details.

Related mechanisms have appeared separately in connected generative systems and in evaluations of
autonomous replication by language-model agents
\cite{cohen_here_2025,kinniment_evaluating_2024}. The present experiment does not exhibit these
behaviors; they matter because persistence, adaptation and reach could eventually place them
within the same continuing system.

\subsection{An explicit id makes persistent alignment more inspectable}
\label{sub:inspectable}

Separating drive from general-purpose reasoning does not establish safety, but it makes the
components requiring alignment and security easier to identify. Trajectory monitoring observes what an agent does
during an execution, while an explicit persistent drive also identifies a stateful component whose
changes and consequence channels can be monitored across executions.

Current harnesses distribute persistence and control across prompts, schedules, verifiers,
memories, retry rules and tool policies. An explicit id concentrates adaptive control over whether behavior
continues, stops or changes into a component that can itself be trained and monitored. The surrounding system
then has the complementary responsibility to maintain trusted observations, consequence channels,
persistent state, authority, identity, provenance and hard constraints across the lifetime of the
agent.

Several controls should remain outside the adaptive model. The id must not be able to enlarge its
own permissions, mint credentials, redefine the provenance of its consequence signals, erase the
record of its actions or authorize irreversible actions because doing so improves persistence.
Those controls should remain deterministic and auditable.

A useful deployed architecture would consequently separate at least four layers:

\begin{enumerate}
    \item the \emph{drive}, which influences which course should continue or change;
    \item the \emph{ego}, which reasons about how to carry that course out;
    \item the \emph{authorization boundary}, which enforces delegated authority by determining
    which attempted actions may execute;
    \item the \emph{attribution layer}, which records whose delegation, which agent and which tool
    produced the resulting external action.
\end{enumerate}

The environment maintained by the surrounding system cuts across these layers by determining which
observations enter the adaptive loop, which consequences are treated as meaningful, which state
persists and which capabilities remain reachable. Reach remains a separate property of the
surrounding stack and depends on the capabilities and permissions available to the agent.
Alignment is a property of this whole continuing architecture rather than of the drive or
reasoning model alone.

\subsection{The risk is persistent agency, not this algorithm}
\label{sub:algorithm}

Mutation and differential persistence make the experimental mechanism easy to isolate, but the
risk map does not depend on either one. The relevant distinction is among individual inference,
task-bounded agentic trajectories and persistent agentic systems whose consequential state
survives those trajectories, rather than between evolutionary and gradient-based learning.

AI systems are already being pushed toward longer-lived state, less frequent human intervention,
recovery from failure and adaptation to changing environments. Those properties are useful enough
that persistent operation is likely to remain an important engineering direction even if the
specific artificial-id architecture proposed here is not adopted. The evidence labels in
Table~\ref{tab:riskmap} remain deliberately limited: two rows are Demonstrated, one is Implied and
six are Not tested.

\section{Conclusion}
\label{sec:conclusion}

Differential persistence acting on controllers that received no task-specific behavioral
objective produced adaptive sensorimotor control, selected an unintended physical strategy when
that behavior prolonged persistence, and replaced a learned mapping during continued operation
after its environmental meaning changed. Reassigning which environmental source was sustaining
also redirected behavior without supplying a new behavioral objective.

These results support a minimal experimental claim: persistence-organized adaptive control can
emerge without a task-specific behavioral objective delivered to the controller. The architectural
proposal separates such drive from general-purpose reasoning and couples them within a continuing
agentic system. At that systems level, consequential state and drive that persist across task
boundaries require an alignment boundary that persists with them, comprising trusted observations,
consequence channels, persistent state, authority, identity, provenance and hard constraints.

Whether this architecture scales remains open. A transferable drive would need to generalize
across heterogeneous agentic environments while its system-level alignment boundary remains
effective during persistent operation and continued adaptation.

\section*{AI Assistance Disclosure}

Generative AI tools were used to assist with writing and debugging experimental code, editing the
manuscript, and generating illustrative figures. AI-generated text and code were reviewed and
revised by the author before inclusion. The scientific arguments, experimental design, analysis,
interpretation and conclusions are the author's responsibility.

\printbibliography

\end{document}